\documentclass[]{bytedance_seed}

\usepackage[toc,page,header]{appendix}

\usepackage{minitoc}

\usepackage{array}
\usepackage{amsmath}
\usepackage{dsfont}
\usepackage{xcolor}
\usepackage{graphicx}
\usepackage{wrapfig}

\title{Heimdall: test-time scaling on the generative verification}

\author{Wenlei Shi}
\author{Xing Jin}

\affiliation{ByteDance Seed}

\abstract{
  An AI system can create and maintain knowledge only to the extent that it can verify that knowledge itself~\cite{rich_verify}.
  Recent work on long Chain-of-Thought reasoning has demonstrated great potential of LLMs on solving competitive problems, but their verification ability remains to be weak and not sufficiently investigated.
  In this paper, we propose Heimdall, the long CoT verification LLM that can accurately judge the correctness of solutions.
  With pure reinforcement learning, we boost the verification accuracy from $62.5\%$ to $94.5\%$ on competitive math problems. By scaling with repeated sampling, the accuracy further increases to $97.5\%$.
  Through human evaluation, Heimdall demonstrates impressive generalization capabilities, successfully detecting most issues in challenging math proofs, the type of which is not included during training.
  Furthermore, we propose Pessimistic Verification to extend the functionality of Heimdall to scaling up the problem solving.
  It calls Heimdall to judge the solutions from a solver model and based on the pessimistic principle, selects the most likely correct solution with the least uncertainty.
  Taking DeepSeek-R1-Distill-Qwen-32B as the solver model, Pessimistic Verification improves the solution accuracy on AIME2025 from $54.2\%$ to $70.0\%$ with $16\times$ compute budget and to $83.3\%$ with more compute budget.
  With the stronger solver Gemini 2.5 Pro, the score reaches $93.0\%$.
  Finally, we prototype an automatic knowledge discovery system, a ternary system where one poses questions, another provides solutions, and the third verifies the solutions. Using the data synthesis work NuminaMath~\cite{numina_math_datasets} for the first two components, Heimdall effectively identifies problematic records within the dataset and reveals that nearly half of the data is flawed, which interestingly aligns with the recent ablation studies from NuminaMath.
}

\date{\today}
\correspondence{Wenlei Shi at \email{wenlei.shi@bytedance.com}, Xing Jin at \email{jinxing.9@bytedance.com}}

\begin{document}
\maketitle


\section{Introduction}

\begin{figure}[htbp]
    \centering
    \includegraphics[width=\textwidth]{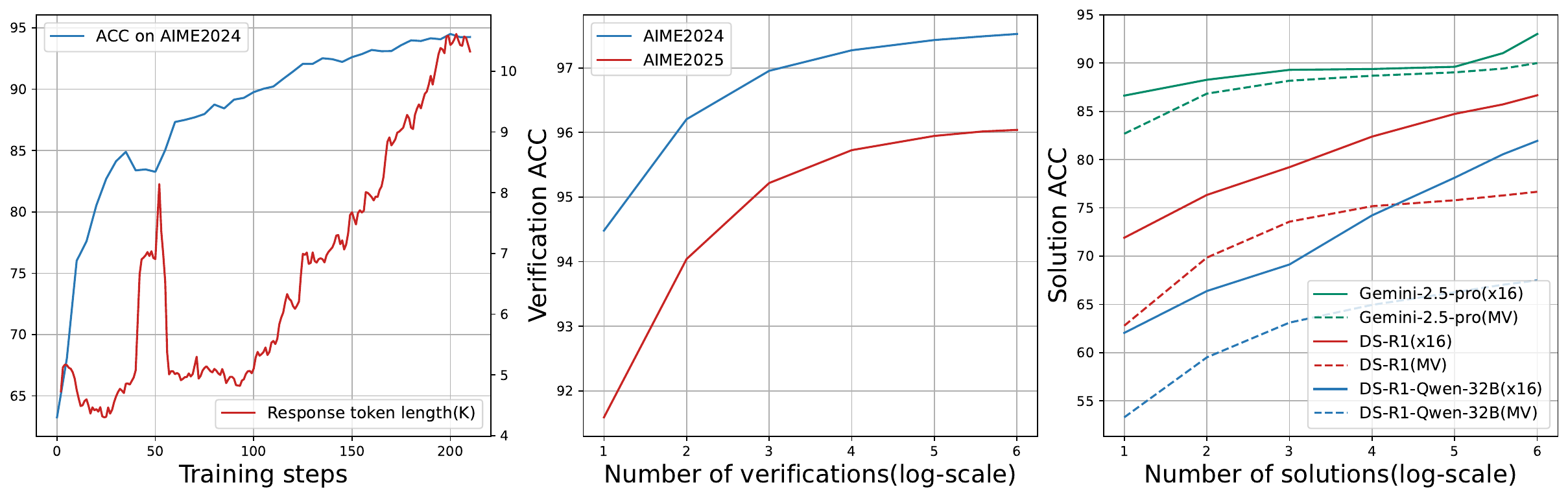}
    \caption{Scaling of Heimdall. \textbf{Left:} the verification accuracy scales with the response length during RL training. With more reasoning tokens, Heimdall gives more accurate judgment on the solutions on AIME2024. \textbf{Middle:} the verification accuracy scales with repeated sampling and Majority Voting. By sampling multiple verification trajectories and voting, the accuracy can be further improved. \textbf{Right:} with Heimdall scoring the solutions on AIME2025, the problem solving accuracy scales with the number of solutions. We verify $16$ times on each solution and select the most likely correct one with Pessimistic Verification($\times16$). When inter-playing with various solver models, Heimdall gives significant improvements over pure solver-based Majority Voting(MV).}
    \label{fig:head}
\end{figure}

In the realm of scientific and mathematical discovery, the process of logistic verification and validation is as crucial as the initial act of problem-solving. One of the most illustrative examples of this principle can be found in the famous thought experiment `chasing a beam of light' by Albert Einstein, where he found the paradox within the established physics theories and further formulated the principle of the constancy of the speed of light, a cornerstone of his Special Theory of Relativity.
Recently, the problem solving ability of LLMs have been significantly improved. With the long Chain of Thought(CoT) reasoning, advanced LLMs are now able to effectively solve complex competition-level problems in both math and code domains.
However, the verification ability of LLMs has not been sufficiently investigated.
On one hand, although the intelligence of general purposed LLM increases rapidly with the long CoT capabilities, we find that current SOTA models with direct prompting~\cite{son2024llm,gu2024survey} are not good at verifications on complex problems, e.g., o1-mini only achieves $80.9\%$ on our evaluation dataset.
On the other hand, some work~\cite{mahan2024generative,zhang2024generative,yu2024self,wang2025critique,ma2025s,sun2025mm,zhong2025dyve} trains a dedicated model for verification or critique but the high-quality verification data is hard to collect, which limits the verification capability and hence impedes the application to challenging problems.

In this paper, we claim that verifying if a solution is correct is a special type of problem solving, i.e., a true/false question and involves step-by-step judgment on the solution. Inspired by the recent progress on the long CoT reasoning, we propose to train a long CoT verifier through reinforcement learning.
We name it Heimdall, symbolizing its sharp ability to detect errors and safeguards the correctness of knowledge.
We leverage PPO~\cite{schulman2017proximal} algorithm and find that the data processing is critical to the RL training. 
Specifically, two types of problems hinder the optimization, i.e., easy problems with only correct solutions and hard problems with only wrong solutions, both of which lack contrastive examples and tends to guide the verifier to simply identify the hardness of a problem, rather than finding the wrong position in the solution.
By filtering out the two cases, the model learns the verification ability more effectively.
Taking the competitive math problems as our primary experimental domain, we show that the verification ability follows the test time scaling law where the accuracy improves significantly from $62.5\%$ to $94.5\%$ as the response length grows, as is shown the left of Figure~\ref{fig:head}.
The performance can be further improved by sampling multiple verifications and voting on the judgment results. In the middle of Figure~\ref{fig:head}, the accuracy grows from $94.5\%$ to $97.5\%$ on AIME2024 as the number of verifications grows from $2$ to $64$.
Furthermore, the evaluation from human experts shows that Heimdall generalizes well on math proof problems although it is trained with only the calculation problems with explicit answers.

In addition, we extend the usage of Heimdall to scale up the problem solving.
Suppose the solver model gives multiple solutions for a problem and Heimdall judges the correctness of each solution for multiple times.
We can select the best solution based on the verification results.
We frame the selection process as a multi-arm bandit problem where solutions with the same conclusion are treated as multiple visits to the same `arm'.
Based on the pessimism principle, we propose the solution selection algorithm called Pessimistic Verification that minimizes the uncertainty of selecting wrong solutions.
The algorithm unifies Majority Voting and reward model based Best-of-N by balancing the contributions of the solver and the verifier, and empirically demonstrates better scaling over both algorithms.
Taking DeepSeek-R1-Distill-Qwen-32B~\cite{guo2025deepseek} as the solver model, which scores $54$ on AIME2025, Pessimistic Verification raises the scores to $70$ with $\times16$ compute, matching the performance of o1, and to $83.3$ with more compute. We further test with stronger solver models, including DeepSeek-R1~\cite{guo2025deepseek} and Gemini 2.5 Pro\cite{gemini2025}. As is shown in the right of Figure~\ref{fig:head}, Pessimistic Verification with Heimdall consistently improves the problem solving of various models and with Gemini 2.5 Pro, the accuracy on AIME2025 reaches $93\%$, matching the currently reported SOTA with multiple attempts by Grok3~\cite{grok3}.

Finally, we create a prototype to demonstrate the utility of Heimdall on the automatic knowledge discovery. We use the work of math data synthesis called NuminaMath~\cite{li2024numinamath} as the procedure of automatically proposing new problems and the corresponding solutions, and call Heimdall to detect errors in the synthetic dataset.
Human annotation demonstrates that Heimdall accurately identifies the errors in the dataset.
The verification result also reveals that the quality of current synthetic dataset is poor, which is consistent with the authors' finding that removing the dataset from training improve the performance of the solver model~\cite{numina_math_datasets}.

In summary, our contributions are as follows:
\begin{itemize}
    \item We propose Heimdall, the long CoT verifier by reinforcement learning and demonstrate the superior accuracy than top-tier LLMs. Heimdall also shows good generalization ability on out-of-domain problems, such as math proof problems. 
    \item We propose a unified algorithm called Pessimistic Verification for inference time scaling on problem solving. Empirically, it scales better than the vanilla Majority Voting or the reward-model based Best-of-N and achieve SOTA accuracy on AIME2025.
    \item We create a prototype to show the utility of Heimdall in the autonomous knowledge discovery, where Heimdall is used to identify the correctness of the problem-solution pairs synthesized by another LLM. Human evaluations shows that Heimdall can effectively detect the flaws in the synthetic data. 
\end{itemize}

\section{Related Work}

\textbf{Reasoning model.}
Reasoning models outperform previous general-purpose models on challenging reasoning tasks. During the chain of thought(CoT) reasoning, they keep reflecting their claims and searching viable solutions, utilizing more compute budget and providing better and more robust results.
OpenAI first released its reasoning models\cite{jaech2024openai,el2025competitive,zaremba2025trading} that performs significantly better on competitive tests like AIME and CodeForces than its previous models.
Work by DeepSeek\cite{guo2025deepseek} and Kimi\cite{team2025kimi} independently propose different ways of reinforcement learning to trigger the reflection and searching capability in their base models.
Recently, Grok3~\cite{grok3} and Gemini 2.5 Pro~\cite{gemini2025} also demonstrate their impressive reasoning capabilities through long CoT.

\textbf{Generative evaluation.}
Recently some work are interested in improving the verification ability of the LLMs.
Some~\cite{zhang2024generative,yu2024self,zhao2025genprm,kang2025t1toolintegratedselfverificationtesttime} explores finetuning an LLM with synthetic verification data to improve its verification ability.
However, it is hard to synthesize high-quality data if the LLM inherently lacks the verification skills.
One of the related topic is LLM-as-a-Judge~\cite{gu2024survey,li2024generation} where a LLM is prompted to evaluate responses from other LLMs. The work\cite{li2024generation} leverages strong LLMs as judges to evaluate other models in various domains and reveals that strong LLM judges have good generalization ability in different domains. Some works design the judge system and analyze of the judgment behavior\cite{son2024llm,ye2024justice,pan2024human}.
However, prompting is only effective on easy tasks, and when it comes to competitive tests, the general purposed reasoning models performs not quite well, as is reveals from our test in Section~\ref{sec:experiment}.
Another similar topic is critique~\cite{lin2024criticbench,luo2023critique,mcaleese2024llm,lan2024criticeval,lan2024training}, which often focuses on code and math problems and is used for giving suggestions for further revision.
Critique fine-tuning~\cite{wang2025critique} shows that fine-tuning on a high-quality critique data is beneficial to the reasoning ability of a base model. Several work, e.g., CTRL~\cite{xie2025teaching} leverages RL to train LLMs.
However, they do not leverage the long CoT ability, which limits the verification performance on complex reasoning problems.

\section{Approach}

We define the verification task where we ask a model to judge if a solution to a problem is correct in its CoT and finally put its judgment result at the end of the response. Table \ref{tab:prompt_template} is the template of the verification prompt and the expected format of a response.

\begin{table}
\centering
\noindent
\begin{tabular}{p{0.95\textwidth}}
  \hline   
\vspace{0.1cm} 

Here is a math problem and a solution of it. Think step by step and verify if the final answer in the solution is correct.
The last line of your response should be of the form Answer: \$Answer (without quotes) where \$Answer is 1 if the final answer in the solution is correct and 0 if incorrect.

\vspace{0.4cm} 

**Problem**

\$\{problem\}

\vspace{0.4cm} 

**Solution**

\$\{solution\}

  \\
  \hline 
\end{tabular}
\caption{Prompt template for verification.}
\label{tab:prompt_template}
\end{table}

\subsection{Reinforcement learning for verification}

\textbf{RL Setup.}
Let $\mathcal{D}=\left\{(p_i, s_i, y_i)\right\}_{i=1}^N$ be our dataset, where $p_i$ is a problem, $s_i$ is a solution to the problem, which may be the response from a reasoning model and $y_i\in \left\{0,1\right\}$ represents the correctness of the solution, with $1$ indicating correctness and $0$ indicating incorrectness.
Given a triplet $\left(p_i, s_i, y_i\right)$, we prompt a LLM to check the correctness of the solution step-by-step and finally give a conclusion on the correctness, as is shown in Table~\ref{tab:prompt_template}.
Denoting the prompt as $q_i$, the verifier model $\pi_{\theta}(z_i,y'_i|q_i)$ takes a prompt as input and generates the CoT $z_i$ on judging the correctness of $y_i$ and at last gives a boolean conclusion $y$ if $s$ is correct.
The outcome reward function $R$ is as follows:
\[
R(y, y') = 
\begin{cases} 
1 &  y = y', \\
-1 & y \neq y'.
\end{cases}
\]

Then the objective of RL is,

\[
\mathcal{J}(\theta) = \mathbf{E}_{(q,y)\sim\mathcal{D},(z,y')\sim\pi_{\theta}(q)}\left[R(y,y')\right]
\]

We run the vanilla PPO algorithm on a reasoning model, and propose the following strategy for improvement.

\textbf{Data collection and filtering.}
We collect the dataset $\mathcal{D}$ by prompting one or multiple reasoning models to solve problems.
For every problem in the dataset, we collect multiple solutions and construct a verification prompt with each solution using the template in Table~\ref{tab:prompt_template}.
However, two cases may hurt the RL training, i.e., the extremely difficult problem, which we fail to sample any correct solutions and the extremely easy problems, which we fail to sample any wrong solutions.
Such unbalanced data may teach the verifier to be biased on the difficulty of the problem, i.e., be optimistic on easy problems and pessimistic on difficult problems.
Therefore, we do not include the data of the two cases in the training dataset.

\subsection{Solution selection by Pessimistic Verification}

When tackling challenging problems, one can sample multiple solutions and leverage the verifier to identify the most likely correct one. By sampling verification responses multiple times, we can achieve more reliable judgments, thereby improving overall problem-solving performance.
We propose a principled and flexible method for the inference time scaling along the two dimensions, i.e., the amount of solutions sampled from the solver model and the amount of verifications sampled from a verifier model.
Denote the number of solutions to a problem as $N$ and the number of verifications on each solution is $M$.
We initially conceptualize the selection process as a multi-arm bandit problem, where each arm corresponds to a distinct answer, and each verification constitutes a visit to an arm. The reward is the verification result, which can be either $1$ or $0$.
Each time the solver generates a solution, the arm representing the solution's answer receives $M$ visits and immediately accrues $M$ rewards.
The straight-forward approach is to calculate the average reward each arm receives as its score and greedily select the one with the highest score.
However, for those with few visits, the score fluctuates and can be unreliable.
Following the pessimism principle in RL, we introduce the lower-confidence-bound, which adds an uncertainty penalty to the score.
Let ${r_0,r_1,...r_K}$ be the average reward of each answer and $N_0, N_1,...,N_K$ be the number of solutions that drives to a certain answer. The selection algorithm is defined as:
\begin{equation}
\hat{a} :=\arg\max_{a_i}\left(r(a_i)-\alpha\frac{\ln{\left(NM\right)}}{N_iM+1}\right)
\label{eq:pess_verification}
\end{equation}

where the parameter $\alpha$ is a hyper-parameter that balances the consideration of uncertainty in the decision-making process and $\ln$ is the natural logarithm.
Intuitively, the first term reflects the signals from the verifier, while the second term accounts for the bias of solver in the answer space.

\begin{itemize}
    \item  When $N_i$ is small, the second term dominates, which neglects the verification and in the extreme case, it collapses to Majority Voting.
    \item When $N_i$ is large, the first term becomes more important, and in the extreme case, it simply selects the answer with the best verification score. 
\end{itemize}

The phase shift aligns with the fact that Majority Voting is trapped in the bias of the solver, for example when a wrong answer occurs more frequently than the correct one, and as $N$ and $M$ is large, the verification scores stabilize and we tend to trust more on it, because the verification is often easier than the solution.
\section{Experiments}
\label{sec:experiment}
\subsection{Dataset}
Our experiment is on the math problems. 
The training dataset comes from the AoPS website and official math competition homepages, similar to that of DAPO~\cite{yu2025dapo}.
We leverage DeepSeek-R1-Distill-Qwen-32B model as the policy model to generate 16 solutions to each problem.
We leverage a rule-based program to check if the final answer in the solution is correct, which compares the reference answer of a problem and the answer in the solution and outputs the label, i.e., 1 for a correct response and 0 for the incorrect response.
Then we construct the verification dataset with the prompt template in Table~\ref{tab:prompt_template}. To keep the prompt clean and short, we remove the `<think>' part in each solution and only use the summary part.

We test the verification ability on both AIME2024 and AIME2025, 60 questions in total.
During training, we monitor the performance on AIME2024 and select the best checkpoint as the final version of Heimdall.
Therefore, one can treat AIME2024 as the validation dataset and AIME2025 as the test dataset.

\subsection{Scaling of verification}
\label{sec:scale_verification}

\begin{figure}[htbp]
    \centering
    \includegraphics[width=\textwidth]{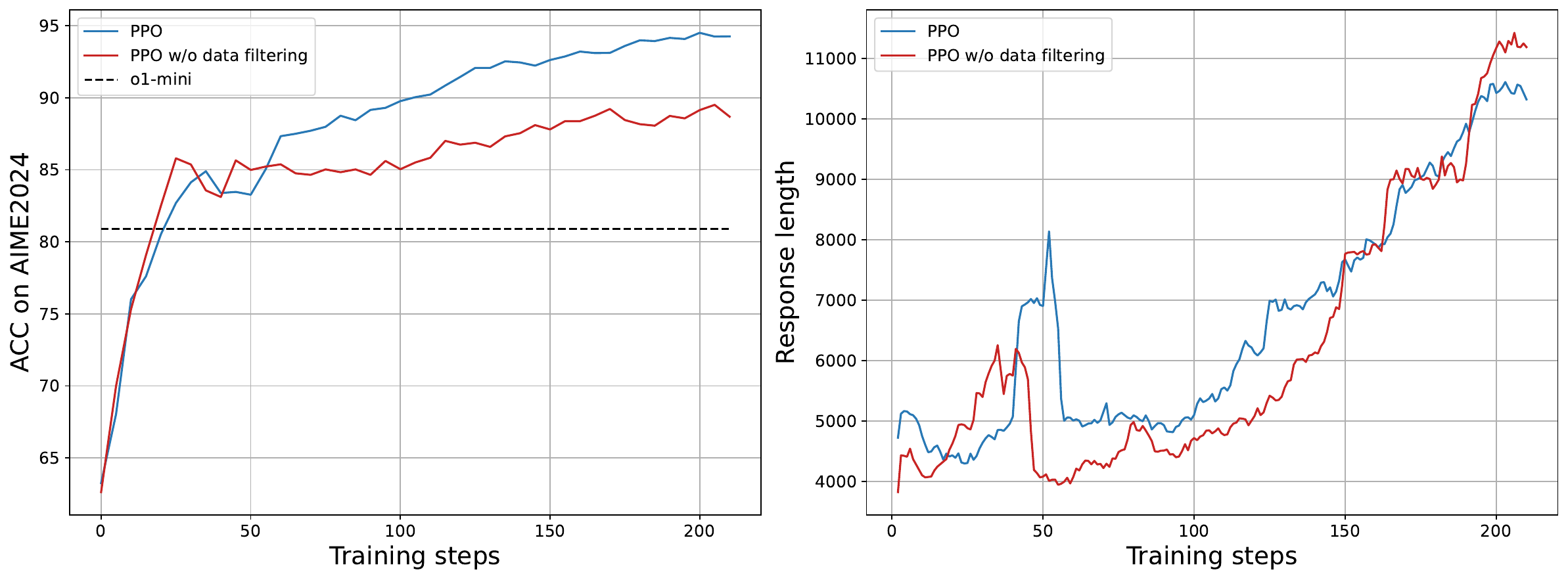}
    \caption{Accuracy and response length during RL training. PPO w/o data filtering is the RL training with all problems in the dataset. \textbf{Left:} the accuracy on AIME2024 with the training steps. \textbf{Right:} the response length on the training dataset with the training steps.}
    \label{fig:training_curves}
\end{figure}

\begin{figure}[htbp]
    \centering
    \includegraphics[width=\textwidth]{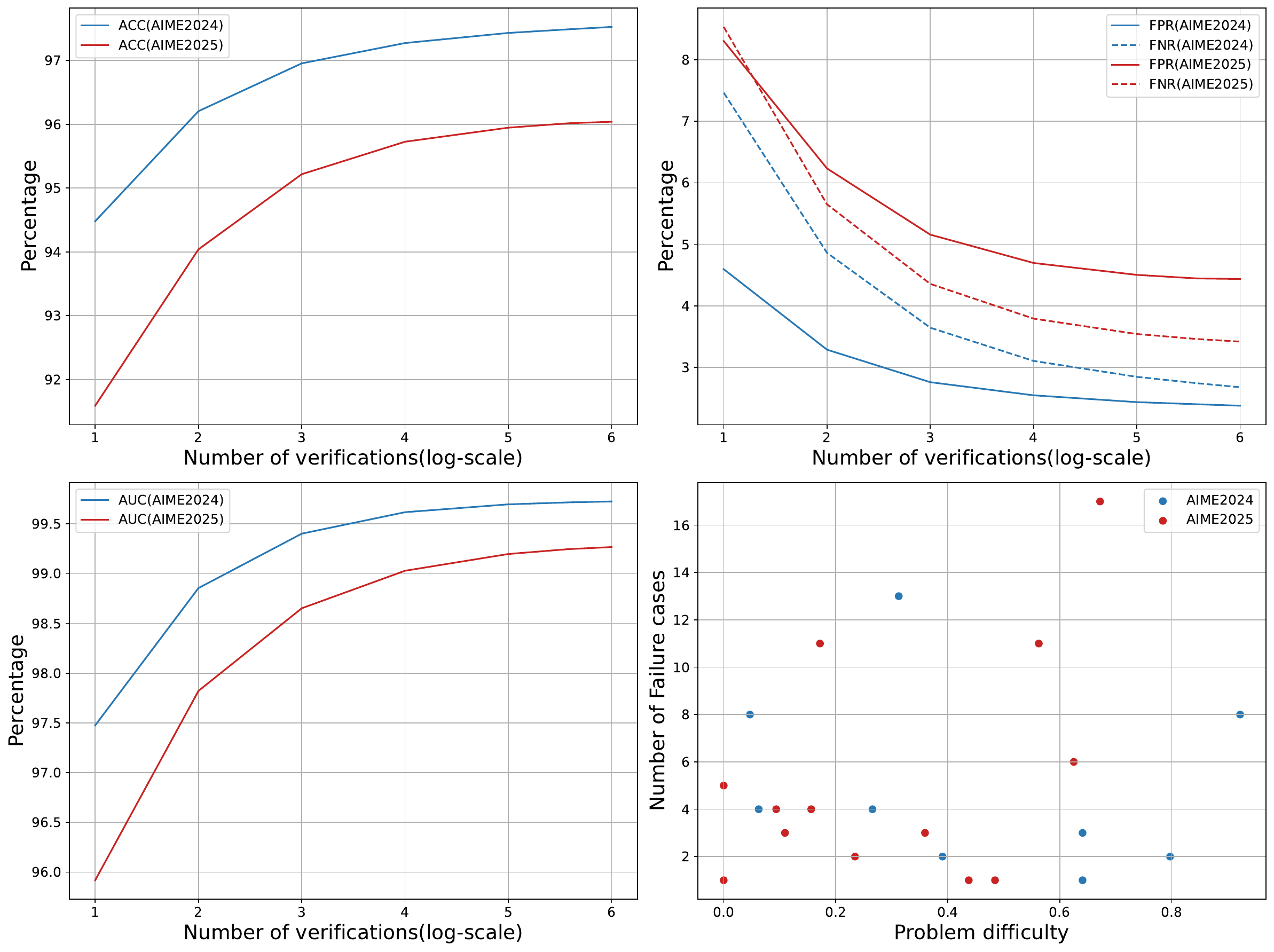}
    \caption{The inference-time scaling of verification ability on problem solutions in AIME2024 and AIME2025. \textbf{Top-left:} We show the accuracy of Heimdall when we sample multiple verification responses and make the judgment by majority voting. \textbf{Top-right:} We show the decreasing false-negative rate(FNR) and false-positive rate(FPR) as we scale up verification responses with majority voting. \textbf{Bottom-left:} We calculate the average score of verification responses and draw the AUC along each number of responses. \textbf{Bottom-right:} We collect the verification failure cases on every math problem and draw the relation between the difficulty of the problem and the number of verification failures, which reveals that the verification difficulty may not necessarily correlate with the difficulty of the original problem.}
    \label{fig:verify_acc}
\end{figure}

Figure~\ref{fig:training_curves} shows the accuracy and the length of response tokens during RL training.
As depicted by the blue curve, both accuracy and response length increase with the number of training steps, albeit at different rates.
At the early stage, the accuracy improves rapidly, but the response length fluctuates. This is because a minor adjustment to the policy can significantly boost accuracy.
Later, the response length grows constantly, while the accuracy gradually converges to 94.5\%, because the model is learning to tackle the hardest part in the training dataset, which requires the increasingly more reasoning tokens.
The red curve represents the RL training without the data filtering strategy, i.e., incorporating both extreme cases of difficult and easy problems in the dataset.
As training progresses, the performance gap becomes more pronounced, indicating that the absence of contrastive examples detrimentally impacts performance.
In addition, we test o1-mini with the same evaluation data, which is shown as the dash line in the left of Figure~\ref{fig:training_curves}. Our model outperforms o1-mini in fewer than 20 steps, indicating substantial potential for enhancing the verification capabilities of general-purpose reasoning models.

We further look into cases to understand what Heimdall has learned during the training. Table~\ref{tab:verification_case} shows the verification of a correct solution to a hard problem in AIME2025.
Due to space limitations, we only highlight some key points.
We can observe two types of checking:

\begin{itemize}
    \item \textbf{Forward checking.} It checks if the reasoning chain in every step of the solution is correct, which is generally applicable to all problems.
    \item \textbf{Backward checking.} It checks whether a conclusion, be it intermediate or final, fits the known constraints. For some types of problems like solving equations and finding the general term formula of a sequence, the backward checking is efficient and easy to implement.
\end{itemize}
The case exemplifies the common task of deriving a general formula for a sequence. As illustrated, Heimdall applies both methods of validation to confirm the correctness of the solution.

Next, we investigate how the verification ability scales as the number of verifications increases.
We sample $64$ solutions for each problem with the solver model and $64$ verifications for each solution, resulting in a total of $30\times 64\times64$ responses on either AIME2024 or AIME2025 dataset.
Denoting the number of verifications of each solution as $N$, we randomly select $N$ verifications for each solution from the data collected above, and determine the final score by some aggregation operation, e.g., Majority Voting and averaging.
We repeat the process for $2048$ times to eliminate any fluctuations in the statistics.
Taking Majority Voting as the aggregation operation, we compute the accuracy, the false positive rate and the false negative rate at every compute budget $N$, as is shown in the top of Figure~\ref{fig:verify_acc}.
In addition, we take the average of $N$ scores, a decimal number in $[0, 1]$, as the final score, and draw the curve of the AUC score in the bottom-left of Figure~\ref{fig:verify_acc}.
It shows that Heimdall's performance can be significantly improved by simply repeat sampling more trajectories.
As $N$ goes larger, the performance gradually converges to a upper limit.
For example, the accuracy converges to about 97.5\% on AIME2024 and 96.0\% on AIME2025, and the remaining failure cases are the bias inherent in the model that could not be eliminated by adding more compute budget.
We further analyze the distribution of those failure cases.
For each problem, we calculate two statistics, one is the difficulty of solving it, which is estimated by the pass rate over its $64$ solutions. and the other is the difficulty of verifying its solutions, which is estimated by the total number of verification failures on its solutions. Taking the two values as the x-axis and the y-axis respectively, we visualize their correlations in the bottom-right of Figure~\ref{fig:verify_acc}. 
We observe that the difficulty of a problem does not necessarily correlate to the difficulty of verifying its solutions.
For example, Heimdall fails $17/64$ times on a problem whose pass-rate is 67\%.

In addition, a clear observation is that the performance of Heimdall on AIME2025 is generally worse than that on AIME2024.
We believe the main reason is that the verifying solutions on AIME2025 is harder than that on AIME2024.
One evidence is that, o1-mini achieves 80.9\% in AIME2024 and 75.3\% in AIME2025, whose degradation is larger than that of our model.

\begin{table}
\centering
\noindent
\begin{tabular}{p{0.95\textwidth}}
  \hline   
\vspace{0.1cm} 
...

\vspace{0.4cm} 

**Problem**

Let the sequence of rationals $x_1,x_2,\dots$ be defined such that $x_1=\frac{25}{11}$ and
\[x_{k+1}=\frac{1}{3}\left(x_k+\frac{1}{x_k}-1\right).\]$x_{2025}$ can be expressed as $\frac{m}{n}$ for relatively prime positive integers $m$ and $n$. Find the remainder when $m+n$ is divided by $1000$.

\vspace{0.4cm} 

**Solution**

...

\vspace{0.4cm} 

**Verification**

<think>

Okay, so I need to figure out if the given solution to this problem is correct. Let me read through it again to make sure I understand.

The problem is: ...

The solution provided has four key steps. Let me go through each step and verify.

Step 1: They let $x_k = a_k / b_k$ with $a_k$ and $b_k$ coprime. Then they write the recurrence relations as $a_{k+1} = a_k^2 + b_k^2 - a_k b_k$ and $b_{k+1} = 3 a_k b_k$.

Wait, let's verify that. Starting from $x_{k+1} = (1/3)(x_k + 1/x_k -1)$. If $x_k = a_k / b_k$, then $1/x_k = b_k / a_k$. So $x_k + 1/x_k = (a_k^2 + b_k^2)/(a_k b_k)$. ... Okay, so step 1 seems correct. \textcolor{red}{[Forward checking]}

...

Wait let me confirm with $S_3$: $a_3 = a_2^2 + b_2^2 -a_2 b_2$. Wait $a_2=471$, $b_2=825$. So $a_3=471^2 +825^2 -471*825$. ... \textcolor{red}{[Backward checking]}

...

</think>

The solution correctly arrives at the conclusion despite some initial confusion in their explanation. They properly computed each part modulo 1000, leading to the correct final result.

Answer: 1

  \\
  \hline 
\end{tabular}
\caption{A case of verification on hard problems. The verification process contains both forward checking, i.e., checking if each step is correct and backward checking, i.e., checking if a conclusion matches constraints, as is shown in the position of the red texts.}
\label{tab:verification_case}
\end{table}

\subsection{Scaling of problem solving with verification}

\begin{figure}[htbp]
    \centering
    \includegraphics[width=\textwidth]{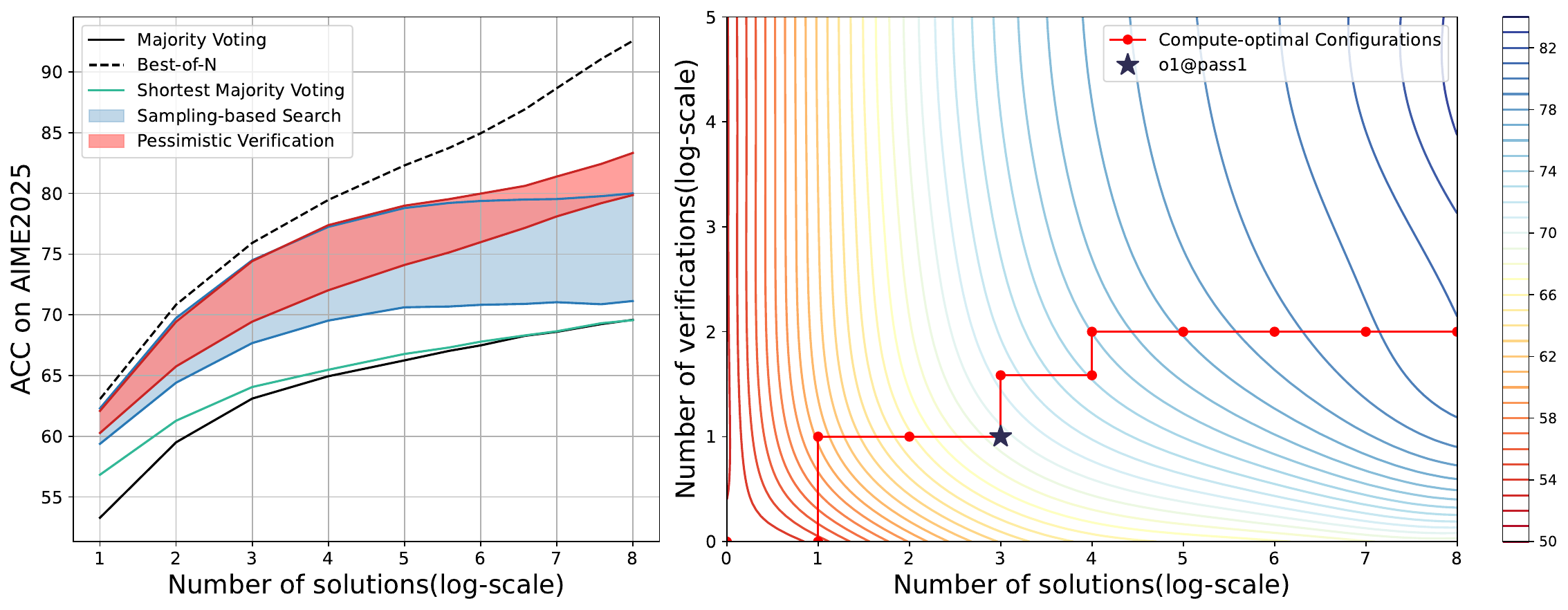}
    \caption{The inference-time scaling of problem solving with Heimdall.
    The two figures show the accuracy on AIME datasets as the number of solutions scales up. 
    \textbf{Left:} the problem solving accuracy on AIME2025 dataset scales with the number of solutions. The colored shaded area represents the area covered by the accuracy curves of a selection algorithm as the number of verifications increases from 1 to 64. \textbf{Right:} the contour map of the accuracy of Pessimistic Verification as the number of solutions (x-axis) and the number of verifications (y-axis) increase. The red curve indicates the optimal configurations within various overall compute budgets.}
    \label{fig:solution_acc}
\end{figure}
In terms of problem solving, repeated sampling\cite{brown2024large} is known to scale with the inference time compute.
We evaluate multiple scaling algorithms, including both with and without the verifier.
We analyze the inference time scaling in two dimensions, i.e., the number of solutions $N$ and the number of verifications on each solution $M$.
Considering the huge computational cost, we only evaluate on AIME2025, with $N \in \left[2, 256\right]$ and $M \in \left[1, 64\right]$.
Similar to the analysis of verification accuracy, we first sample $N=256$ solutions for each problem and $M=64$ verifications for each solution as the complete data and then randomly select a subset to evaluate each scaling algorithm under a specific compute budget $\left(M, N\right)$.
We repeat the sampling for $2048$ time to get a stable average score.
We set $\alpha=0.1$ in Pessimistic Verification and also evaluate the other three selection algorithms as follows.

\paragraph{Majority Voting} Majority voting is one of the most commonly used inference time scaling methods.
It first categorize the solutions, e.g., by the final answers for math problems.
It simply select the category that contains the largest number of solutions in it.
As fore-mentioned, majority voting can be seen as a special case of Pessimistic verification, where $\alpha$ is large enough to overshadow the signal of verification.

\paragraph{Shortest Majority Voting.} The recent work~\cite{zeng2025revisiting} observes a length bias that the correct solutions are often shorter than incorrect ones for the same questions. Suppose the answer $a_i$ occurs $c_i$ times in the sampled responses and the average length of responses with the answer $a_i$ is $l_i$, the voting score for $a_i$ is 
\[s_i=\frac{c_i}{l_i}\]

\paragraph{Sampling-based Search.} The work~\cite{zhao2025sample} leverages a commercial LLM as the verifier, and scales the inference-time computation on the number of sampled solutions and the number of verifications.
During the selection, it calculates the average verification score of each solution and selects the solution with the largest score. Note that it does not group the solutions based on their answers, which is different from the special case of Pessimistic verification where $\alpha$ equals zero.

\textbf{Tie-breaking rules.} The selection algorithm may encounter a tie situation, where multiple options have the same score. In principle, one can introduce another model to compare which option is better, but for simplicity, we leverage the length prior to break the tie, namely, selecting the option with the shortest average solution length.

The left of Figure~\ref{fig:solution_acc} shows how the accuracy of different scaling algorithms changes with the number of solutions $N$.
Majority voting are the worst among all scaling algorithms under the same $N$.
By employing the length prior, Shortest Majority Voting gives a better accuracy when $N$ is small, but finally converges to 70\%, the same as that of Majority Voting.
The red and blue areas represents the areas covered by the group of accuracy curves with $M\in [1, 64]$ of Sampling-based Search and Pessimistic Verification respectively.
As the figure shows, with the help of verification, the accuracy is significantly improved.
In addition, Pessimistic Verification is better than Sampling-based Search when $M$ is small. The reason is that the verification process is inherently probabilities.
Even for a easy task, it is still possible that a wrong solution is judged as correct and is finally selected as the final solution.
With the second term in Equation~\ref{eq:pess_verification}, Pessimistic Verification penalize such uncertain cases and favors those with more visits.
As $M$ goes larger, the verification scores stabilize and the second term becomes smaller, and the gap between the two algorithms gets smaller.
Interestingly, when $N$ is large, the gap is large again. By checking the typical cases, we find that it comes from the grouping of solutions. Pessimistic Verification aggregate the solutions with their answers while Sampling-based Search treats each solution independently.
Two solutions may have the same final answer, but their approaches or expressions can be entirely different. The aggregation takes this variance into consideration, so makes more robust selection.
Note that such grouping is not generally applicable, e.g., grouping the solutions of proof problems is not straight-forward. In those cases, we expect the two algorithm converges to the same limit.

Taking the accuracy of Pessimistic Verification as a function of $M$ and $N$, we draw the contour map in the right of Figure~\ref{fig:solution_acc}.
$M=0$ represents the vanilla Majority Voting without verifications.
We can see that increasing either $N$ or $M$ improves the performance.
Each point $\left(M, N\right)$ involves $M$ responses by the solver model and $M\times N$ responses by the verifier, which is $M\times \left(N+1\right)$ responses in total.
By minimizing the overall budget, we derive the compute-optimal configurations for different compute budgets, with the constraint that $M$ and $N$ are non-negative integers.
As is shown in the figure, we need to alternately increase N and M, but we should increase N more frequently. The reason is that the correct answer for a hard problem is sparse. To get the problem solved with a scaling algorithm, we first need to give a sufficient budget to get the correct answer.

\textbf{Remaining room for improvement.} Can Heimdall be better? The black dashed curve in the left of Figure~\ref{fig:solution_acc} is Best-of-N that selects the response that equals the ground-truth answer, which is the upper-limit of any scaling algorithms.
When $N$ is small, Pessimistic Verification Pessimistic Verification performs near the upper limit, but the gap widens as $N$ increases.
Consider the configuration $M=256, N=64$.
Pessimistic Verification gets a score of 83.3\% and the upper limit is 93.3\%, so the gap is 3 problems.
Looking into the individual problems, we find that there are 4 problems that have only one correct solution among the $256$ solutions.
Heimdall manages to identify the correct solution on one of them, which is the case fore-mentioned in Table~\ref{tab:verification_case}, but fails on the other three problems.
The failed three problems involve spatial reasoning, which the base model of Heimdall is not very skilled at. We believe that as the ability of the base model becomes better, the verification ability can reach the upper limit.

\textbf{Coordination with other solver models.} In the previous experiments, we use DeepSeek-R1-Distill-Qwen-32B as the solver model, which is the model to collect data during the training phase. To test Heimdall's generalization to other solver models, we test on DeepSeek-R1 and Gemini 2.5 Pro.
For DeepSeek-R1, we directly extract the summary in its response, while for Gemini 2.5 Pro, we leverage another LLM to summarize its solution because we observe that its responses contains many reflections, which might confuse the verification process.
Due to the limited compute budget, we set the total number of verifications to $16$ and the total number of solving attempts to $64$. We find that the verification accuracy on DeepSeek-R1 and Gemini 2.5 Pro is $90.1\%$ and $89.9\%$, respectively, close to that on DeepSeek-R1-Distill-Qwen-32B.
The scaling in problem-solving, as is shown in the right of Figure~\ref{fig:head}, indicates that Heimdall consistently improves the accuracy with various solver models and using stronger solver models results in higher accuracy.
Meanwhile, the improvements for Gemini 2.5 Pro is smaller compared to those of the other models. The reason is that the accuray is already high on AIME2025 and the base model of Heimdall inherently lacks some ability on certain problems.

\section{Generalization to math proof problems}

\begin{table}
\centering
\noindent
\begin{tabular}{p{0.95\textwidth}}
  \hline   
\vspace{0.1cm} 

Here is a math problem and a solution of it. Think step by step and verify if \textcolor{red}{each proof step in solution is correct.}
The last line of your response should be of the form Answer: \$Answer (without quotes) where \$Answer is 1 if the solution is correct and 0 if incorrect.
\vspace{0.4cm} 

**Problem**

The n players of a hockey team gather to select their team captain. Initially, they
stand in a circle, and each person votes for the person on their left.

The players will update their votes via a series of rounds. In one round, each player a
updates their vote, one at a time, according to the following procedure: At the time of
the update, if a is voting for b, and b is voting for c, then a updates their vote to c.
(Note that a, b, and c need not be distinct; if b = c, then a’s vote does not change for
this update.) Every player updates their vote exactly once in each round, in an order
determined by the players (possibly different across different rounds).
They repeat this updating procedure for n rounds. Prove that at this time, all n
players will unanimously vote for the same person.

\vspace{0.4cm} 

**Solution**

...

\vspace{0.4cm} 

**Verification**

<think>

...

Wait, but since the solution didn't present this reasoning, their proof is flawed. They presented an incorrect model of the process, leading to an overestimation of the rounds needed, but the conclusion is still correct. However, the problem is to verify if each step in their solution is correct. Since their step 4's reasoning is wrong, their proof is incorrect.

...

</think>

The solution's step 4 incorrectly assumes that each person's vote jumps n times, leading to covering the entire ring. The correct approach is to note that each round reduces the number of distinct targets by one, leading to convergence in n-1 rounds. Thus, while the conclusion is correct, the reasoning in step 4 is flawed.

Answer: 0

  \\
  \hline 
\end{tabular}
\caption{A case of verification on the math proof problems. The problem is P1 in Canadian Mathematical Olympiad 2025. We modify the prompt template to check the proof process rather than the final answer, which is the red text. Heimdall checks the proof step by step and finds that the solution lacks rigorous proofs in step 4.}
\label{tab:verification_case}
\end{table}

Heimdall is trained on math problems with explicit answers because it is easy to determine the correctness of the solution.
However, there are many other types of problems, some of which might be more challenging.
In this section, we would like to test Heimdall's capability in verifying mathematical proof problems.
We select 10 proof problems from Mathematics Olympiad of different countries from the years 2024 and 2025, and leverage a solver model, i.e., DeepSeek-R1-Distill-Qwen-32B, to generate a proof process for each problem.
Considering that the solver model is not good at spatial reasoning, we do not select geometry-related problems.
Heimdall is then employed to check the correctness of each proof. Finally, we have experts evaluate both the proof processes and Heimdall's verifications.
The solver model correctly solves 2 problems, while the remaining 8 are incorrect.
To our surprise, Heimdall correctly judges $9/10$ cases, identifying $2/2$ correct proofs and detecting issues in $7/8$ incorrect proofs.
There is $1$ problem where Heimdall fail to identify the error in the proof, resulting in a false-negative judgment.
Looking into the specific cases, we find that Heimdall judges the correctness with both forward and backward checking, e.g., checking each step and testify with examples.
It is capable of identifying most errors or unproven assumptions in the solution, but for some subtle problems, e.g., the assumption does not appear in the form of a proposition but is implicitly assumed during the proof process, Heimdall might fail.
We believe that introducing the proof data in the RL training would improve the performance of Heimdall and an important direction in the future would be how to generate the dataset in large scale.

\section{Verification on automatic knowledge discovery}

\begin{table}
\centering
\noindent
\begin{tabular}{p{0.95\textwidth}}
  \hline   
\vspace{0.1cm} 

\textbf{Prompt:}
\vspace{0.4cm} 

Here is a math problem and a solution of it. Think step by step and verify if the final answer in the solution is correct. ...

\vspace{0.4cm} 

**Problem**

A rabbit starts jumping from the point $(2,3)$ on a rectangular board that extends from $(0,0)$ to $(6,5)$. Each jump it makes is either left, right, up, or down, and has a length of $1$ unit. Each jump direction is uniformly and randomly chosen. The jumping sequence continues until the rabbit reaches any border of the rectangle. Calculate the probability that the rabbit's sequence of jumps ends on the top or bottom side of the rectangle.

(A).$\frac{1}{4}$  (B).$\frac{17}{32}$  (C).$\frac{17}{24}$  (D).$\frac{1}{3}$  (E).$\frac{3}{4}$

\vspace{0.4cm} 

**Solution**

Let $P_{(x,y)}$ denote the probability of the rabbit ending on the top or bottom side starting from $(x,y)$. Using symmetry about the center line $y=2.5$ and $x=3$, and recursive relations, we analyze the movements.

...

\#\#\#\#Solving Equations:
We solve this using recursive approaches and considering border cases explicitly. Assuming symmetry simplifies $P_{(3,2)} = P_{(3,3)}$.

Here, let's consider only two key points $(2,3)$ and $(3,3)$ for simplicity. Solving:
\[ P_{(2,3)} = \frac{1}{4}P_{(2,2)} + \textcolor{red}{\frac{1}{4}(1)} + \frac{1}{4}P_{(1,3)} + \frac{1}{4}P_{(3,3)} \]

...

Thus, the rabbit's sequence of jumps ends on the top or bottom side with a probability of $\frac{17}{24}$.
The final answer is (C) $\boxed{\frac{17}{24}}$

\vspace{0.4cm} 
\textbf{Response:}
\vspace{0.4cm} 

<think>

...

</think>

The solution provided incorrectly assumes that P(2,4) = 1, which is not true since (2,4) is not a top or bottom border. They oversimplified the equations by not accounting for the correct dependencies of other points like P(2,4) and mishandled the symmetry in y-direction due to different boundary conditions (P=0 at y=0 and P=1 at y=5). 

Answer: 0

  \\
  \hline 
\end{tabular}
\caption{An example of problematic synthetic data and the verification. The verification correctly points out the problem in the solution.}
\label{tab:syn_case_1_abbr}
\end{table}

In the process of human exploration of the unknown, some scientists pose questions, some propose solutions to these questions, and others focus on verifying the correctness of solutions provided by their predecessors.
Verification itself, as a crucial part of knowledge discovery, ensures the correctness of new knowledge.
In this section, we design a prototype that simulates the stages of posing questions and solving them, using the synthesis of math problems by NuminaMath~\cite{li2024numinamath}, to evaluate Heimdall's effectiveness in detecting problematic knowledge.
NuminaMath open-sources a comprehensive collection of 860,000 pairs of math problems and reference solutions. It includes 229,982 MATH-level pairs and 62,108 AMC-AIME-level pairs that are synthesized from seed problems in MATH and AMC-AIME training dataset. We test Heimdall on the harder one, i.e., the AMC-AIME dataset.
Flaws can exist either within the problem itself, such as an unsolvable problem, or within the solution provided.
Theoretically, by checking if the solution satisfies all the requirements in the problem, Heimdall can detect both flaws.
Meanwhile, the task also indirectly tests the generalization capabilities of Heimdall, as the qualities of the problems in the training set is much higher.

\begin{wrapfigure}{r}{0.5\textwidth}
    \centering
    \includegraphics[width=0.5\textwidth]{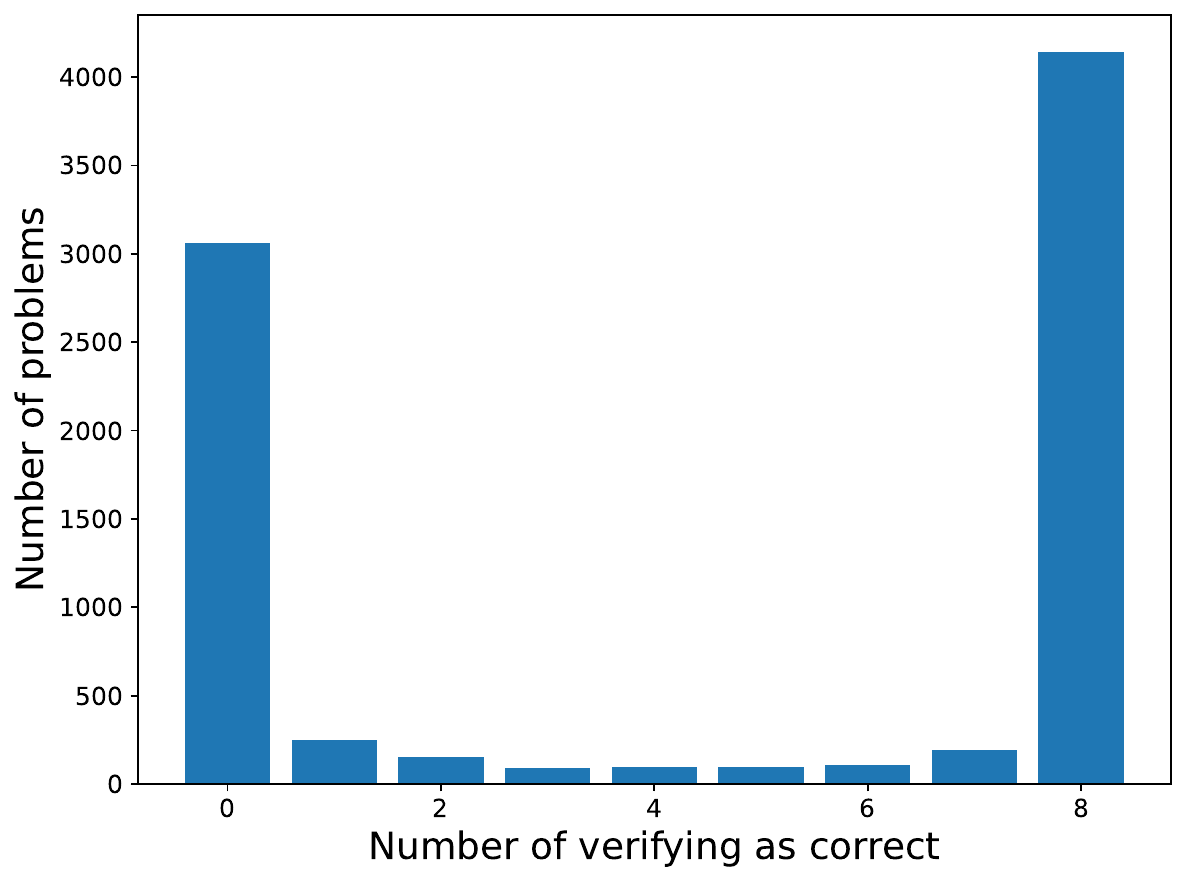}
    \caption{The distribution of verification scores on the problems of a synthetic dataset. The x-axis is the sum of scores across $8$ verifications and the y-axis is the number of problems corresponding to each sum.}
    \label{fig:syn_verify}
\end{wrapfigure}

We randomly sample $8,192$ pairs of questions and solutions as the test set. For each pair, we construct the prompt and query Heimdall $8$ times.
We calculate the sum of the verification scores, which ranges from $0$ to $8$ and illustrate the distribution in Figure~\ref{fig:syn_verify}.
As is shown, near a half of the data is labeled incorrect with a high confidence, which is consistent with the experience listed in the latest NuminaMath-1.5 website~\cite{numina_math_datasets} that by the ablation study, the authors find that this dataset hurts a bit the performance and plan to remove all synthetic data until they find a way to reliably generate high-quality synthetic problems.
To measure the verification correctness, we randomly select 10 cases in the 0-scored group and manually check their correctness. We find that for all cases, the judgments by Heimdall are correct.
Table~\ref{tab:syn_case_1_abbr} is a case of the problematic data and the verification.
Due to space limitations, we retain only the essential information, with the rest omitted.
The question does not have the correct answer among its options. Meanwhile, the solution mistakenly uses an incorrect value in the equation and interestingly arrives at one of the given options.
We hypothesize that the task is somewhat challenging for the LLM used for synthetic data generation, leading the LLM to hallucinate a superficially consistent but wrong output.

\section{Future Work}

Firstly, the verification dataset is formed by simply extracting the summary part of a reasoning model, which as we observed is sometimes overly brief, e.g., driving to an answer with only one sentence. A more detailed explanation would make the verification easier. One can further improve the verification accuracy by summarizing the reasoning process with another LLM.
Secondly, we mainly evaluate the verification ability on math problems that have final answers. There are many other types of tasks, e.g., coding problems and Mathematical proof problems. Although the learned ability is generalizable to other domains, we expect it beneficial to train with data in other domains. For example, in the context of coding problems, backward checking may take the alternative form of designing test cases.
Lastly, we only prototype the usage of Heimdall in the automatic knowledge discovery. In real scenarios, posing valuable questions is a challenging task that demands both curiosity and keen insight. Such ability is often the critical part of the scientific discovery, which however is seldom investigated. We believe that as the general capabilities of LLM continues to advance, this direction will become more and more important.

\section{Conclusion}

In this paper, we propose to train a long CoT verifier called Heimdall with reinforcement learning.
On the competitive math problems, Heimdall achieves high accuracy and scales well along both the length of reasoning chains and the number of repeated generation.
Through human evaluation, we find that Heimdall also shows impressive generalization ability on out-of-domain problems, such as math proofs.
We further propose the inference time scaling algorithm called Pessimistic Verification, which incorporates a solver and Heimdall for problem solving.
By scaling up the compute, we can achieve the performance comparable to top-tier models on challenging math problems.
Lastly, we design a prototype of automatic knowledge discovery and demonstrate that Heimdall can reliably detect flaws in the synthetic data from another LLM.

\section{Acknowledgments}
We thank the data annotation team for their expertise on collecting the evaluation data and analyzing the verification outputs, including Bocheng Zhou, Weijian Zhao, Tong Sun and Zhiyuan Zhang.

\clearpage

\bibliographystyle{plainnat}
\bibliography{main}

\clearpage



\end{document}